\documentclass[10pt, a4paper]{article}
\usepackage{lrec}
\usepackage{graphicx}
\usepackage{tabularx}
\usepackage{soul}

\usepackage{url}

\usepackage{booktabs}
\usepackage{multirow}

\usepackage{amsmath}

\usepackage[linesnumbered,ruled]{algorithm2e}

\SetKwInput{KwInput}{Input}
\SetKwInput{KwOutput}{Output}

\newcommand{\ACRO}[1]{\textsc{#1}}
\newcommand{\ARRAU}{\ACRO{arrau}}

\newcommand{\BERT}{\ACRO{bert}}
\newcommand{\ELMO}{\ACRO{elmo}}

\newcommand{\CNN}{\ACRO{cnn}}
\newcommand{\CRF}{\ACRO{crf}}
\newcommand{\CONLL}{\ACRO{conll}}
\newcommand{\CRAC}{\ACRO{crac}}

\newcommand{\GENIA}{\ACRO{genia}}
\newcommand{\GENIATT}{\ACRO{genia90}}
\newcommand{\GENIATDT}{\ACRO{genia81}}
\newcommand{\LSTM}{\ACRO{lstm}}
\newcommand{\MD}{\ACRO{md}}

\newcommand{\NER}{\ACRO{ner}}

\newcommand{\NP}{\ACRO{np}}

\newcommand{\LEEMD}{\ACRO{Lee md}}
\newcommand{\BIAFFINEMD}{\ACRO{Biaffine md}}
\newcommand{\BERTMD}{\ACRO{Bert md}}
\newcommand{\TABLEEMD}{\ACRO{Lee}}
\newcommand{\TABBIAFFINEMD}{\ACRO{Bia}}
\newcommand{\TABBERTMD}{\ACRO{Ber}}
\newcommand{\HIGHRECALL}{\ACRO{high recall}}
\newcommand{\HIGHF}{\ACRO{high f1}}
\newcommand{\BALANCE}{\ACRO{balance}}

\title{Neural Mention Detection}

\name{Juntao Yu,$^1$ Bernd Bohnet,$^2$ Massimo Poesio$^1$}

\address{$^1$Queen Mary University of London, $^2$Google Research \\
         juntao.yu@qmul.ac.uk, bohnetbd@google.com, m.poesio@qmul.ac.uk\\
         }

\abstract{
  Mention detection is an important 
  preprocessing step %task 
  for
  annotation  and interpretation  
  in applications such as {\NER} and
  coreference resolution, 
  but few stand-alone neural models have been proposed able to handle the full range of mentions.
  In this work, we propose and compare three neural network-based approaches to mention detection. 
  The first approach is based on the mention detection part of a state of the art coreference resolution system; 
  the second %approach 
  uses {\ELMO} embeddings together with a bidirectional {\LSTM} and a biaffine classifier; 
  the third approach uses the recently introduced {\BERT} model. 
  Our best model (using a biaffine classifier) achieves %large 
  gains of up to 1.8 percentage points on mention recall when compared with a strong baseline in a {\HIGHRECALL} coreference annotation setting. 
  The same model achieves improvements of up to 5.3 and 6.2 p.p. when compared with the best-reported mention detection F1 on the {\CONLL} and {\CRAC} coreference data sets respectively in a {\HIGHF} annotation setting.  
  We then evaluate our models for coreference resolution by using mentions predicted by our best model in  start-of-the-art coreference systems. 
  The enhanced model achieved %large 
  absolute improvements of up to 1.7  and 0.7 p.p. when compared with our strong baseline systems (pipeline system and end-to-end system) respectively. 
  For nested {\NER},
  the evaluation of our model on the {\GENIA} corpora shows that our model 
  matches or outperforms state-of-the-art models
  despite not being specifically designed for this task.
  \newline  }

\begin{document}

\maketitleabstract

\section{Introduction}

Mention detection ({\MD}) is 
the task of identifying mentions of entities in 
text. 
It is an important preprocessing step for downstream applications such as 
nested named entity recognition \cite{zheng2019boundary} or
coreference resolution \cite{poesio-stuckardt-versley:book}; 
thus, %As such,
the quality of mention detection affects
both the performance of models for such applications
and the quality of annotated data used to train them
%In addition, it also plays an important rule in many annotation tasks
\cite{phrasedetectivescorpus,poesio2019naaclpd20}. 

Much mention detection research for {\NER} has concentrated on
%Many downstream applications use 
a simplified version of 
%the 
{\MD} that focuses on  proper names only 
(i.e., it doesn't consider as mentions nominals such as \textit{the protein} or pronouns such as \textit{it}), 
and ignores the fact that mentions may nest (e.g., noun phrases such as  \textit{[[CCITA] mRNA]} in the \ACRO{genia} corpus are mentions of two separate entities, \textit{CCITA} and \textit{CCITA and mRNA} \cite{alex-et-al:BIONLP07}).
%({\NER}) 
However such simplified view of mentions
%the {\NER} system 
%alone 
is not sufficient for {\NER} in domains such as biomedical, or for 
%more complex downstream applications such as
coreference, that requires full mention detection.
%Comparing to {\NER}, 
% The full {\MD} task 
% %for coreference resolution 
% is more complex in three
% respects: firstly, it identifies more mention types, such as nominal mentions and pronouns; 
% secondly, the mentions can be nested, so the task cannot be treated as a simple sequence labelling task, as is the norm in {\NER} systems; 
Another limitation of 
%thirdly, 
typical mention detection %{\NER} 
systems 
is that they only predict mentions in a {\HIGHF} fashion, whereas in  coreference, for instance, mentions are usually predicted in a {\HIGHRECALL} setting, since further pruning will be carried out at the coreference system \cite{clark2016improving,clark2016deep,lee2017end,lee2018higher,kantor2019bertee}.%\MPCOMMENT{MP}{We need a reference}.

%The most recent
% \MPCOMMENT{MP}{I would remove the following para as it's a bit obsolete now - these days most coreference systems are end-to-end}
% Neural network approaches using  context-sensitive embeddings 
% such as \ACRO{elmo} \cite{peters2018elmo} and {\BERT} \cite{devlin2019bert} have resulted in substantial improvements 
% %shown their advances 
% in the {\NER} benchmark {\CONLL} 2003 data set.
% %and pushed the performance further up. 
% However, most %of the 
% {\MD} systems used by the state-of-the-art coreference systems %are 
% do not take advantage of these advances and 
% still heavily rely on %the 
% parse trees \cite{bjorkelund2014learning,wiseman2015learning,wiseman2016learning,clark2016deep,clark2016improving}. 
% They either use all the {\NP}s as candidate mentions \cite{bjorkelund2014learning,wiseman2015learning,wiseman2016learning} or use the rule-based mention detector from the Stanford deterministic system \cite{lee-et-al:CL13} to extract mentions from {\NP}s, named entity mentions and pronouns \cite{clark2015entity,clark2016improving}. 

% To build a high accuracy standalone {\MD} system is not only important for  downstream applications, but also beneficial for annotation tasks that require mentions \cite{phrasedetectivescorpus,poesio2019naaclpd20}, 
% but
There are only very few recent studies that attempt to apply neural network approaches to develop a standalone mention detector. 
Neural network approaches using  context-sensitive embeddings such as {\ELMO} \cite{peters2018elmo} and {\BERT} \cite{devlin2019bert} have resulted in substantial improvements 
for mention detectors in the {\NER} benchmark {\CONLL} 2003 data set.
% %and pushed the performance further up. 
%{\MD} task. 
However, most  coreference systems that appeared after Lee et al., \shortcite{lee2017end,lee2018higher} 
carry out mention detection 
%first introduced a neural mention detector 
as a part of their end-to-end coreference system. %however, 
Such systems do %the system does 
not output intermediate mentions, hence the mention detector cannot be directly used 
to extract mentions for an annotation project, 
or by other coreference systems. % directly. 
Thus the only standalone mention detectors that can be used as preprocessing for a coreference annotation 
are ones that do not take advantage of these advances and still heavily rely on parsing  
to identify all  {\NP}s as candidate mentions \cite{bjorkelund2014learning,wiseman2015learning,wiseman2016learning} or 
ones that use the rule-based mention detector from the Stanford deterministic system \cite{lee-et-al:CL13} to extract mentions from {\NP}s, named entity mentions and pronouns \cite{clark2015entity,clark2016improving}. 
To the best of our knowledge, \newcite{poesio2018crac} introduced the only standalone neural mention detector. By using a modified version of the {\NER} system of \newcite{glample2016-ner}, they showed 
substantial performance gains at mention detection on the benchmark {\CONLL} 2012 data set and on the {\CRAC} 2018 data set when compared with the Stanford deterministic system. % \cite{lee-et-al:CL13}. 
% To build a high accuracy standalone {\MD} system is not only important for the downstream applications, but also beneficial for annotation tasks that require mentions \cite{phrasedetectivescorpus,poesio2019naaclpd20}. 

In this paper, we compare three neural architectures for standalone {\MD}. 
The first system is a slightly modified version of the mention detection part of the \newcite{lee2018higher} system. 
The second system employs a bi-directional {\LSTM} on the sentence level and uses 
biaffine attention \cite{dozat-and-manning2017-parser} over the {\LSTM} outputs to predict the mentions. 
The third system takes the outputs from {\BERT} \cite{devlin2019bert}
and feeds them into a feed-forward neural network to classify candidates into mentions and 
non mentions. 
All three systems have the options to output mentions in {\HIGHRECALL} or {\HIGHF} settings;
the former is well suited for the coreference task,
whereas the latter can be used as a standard mention detector for tasks like nested named entity recognition.
We evaluate
our models
on the {\CONLL} and the {\CRAC} data sets for coreference mention detection, and on {\GENIA} corpora for nested {\NER}.

The contributions of this paper are therefore as follows.
%we have the following findings: 
First, we show that 
 mention detection performance improved by  up to 1.5 percentage points\footnote{This performance difference is measured on mention recall, as we follow \newcite{lee2018higher} to use fixed mention/token ratio to compare 
the mentions selected by their joint system.} can be achieved by training the mention detector alone.
Second, our best system achieves 
improvements of 5.3 and 6.2 percentage points when compared with \newcite{poesio2018crac}'s neural {\MD} system on {\CONLL} and {\CRAC} respectively.
Third, by using better mentions from our mention detector, we can improve the end-to-end \newcite{lee2018higher} system and the \newcite{clark2016deep} pipeline system by up to 0.7\% and 1.7\% respectively.
Fourth, we show that the state-of-the-art result on nested {\NER} in the \ACRO{genia} corpus can be achieved by our best model.
%, despite our model is not specifically designed for this task.

\section{Related Work}

\textbf{Mention detection.} 

Despite neural networks 
having shown
high performance
in many natural language processing tasks, the rule-based mention detector of the Stanford deterministic system \cite{lee-et-al:CL13} 
remained %remains 
frequently used in %the 
top performing coreference systems that preceded the development of end-to-end architectures \cite{clark2015entity,clark2016deep,clark2016improving}, including the best neural network coreference
system %that 
based on a pipeline architecture, \cite{clark2016deep}. 
The Stanford Core mention detector uses 
a set of predefined heuristic rules %are used 
to select mentions from {\NP}s, pronouns and named entity mentions. 
Many other coreference systems simply use all the {\NP}s as the candidate mentions \cite{bjorkelund2014learning,wiseman2015learning,wiseman2016learning}. 

\newcite{lee2017end} first introduced a neural network based end-to-end coreference system 
in which the 
neural mention detection part is not separated.
This 
strategy led to a greatly improved performance on the coreference task; 
%move proved very effective;
however, %as a result 
the mention detection component of their system needs to be trained jointly with the coreference resolution part, hence can not be used separately for  applications other than coreference.
The Lee et al system has been later extended by \newcite{zhang2018acl}, \newcite{lee2018higher} and \newcite{kantor2019bertee}. 
\newcite{zhang2018acl} added 
biaffine attention to the coreference part of the \newcite{lee2017end} system, 
improving the system by 0.6\%. 
(Biaffine attention is also used in one of our approaches (\BIAFFINEMD), but in a totally different manner, i.e. we use
biaffine attention for mention detection 
while 
in \newcite{zhang2018acl}
biaffine attention was used for computing mention-pair scores.)  
In \newcite{lee2018higher} and \newcite{kantor2019bertee}, the \newcite{lee2017end} model is 
substantially improved through the use of 
{\ELMO} \cite{peters2018elmo} and {\BERT} \cite{kantor2019bertee} embeddings.

Other machine learning based mention detectors include \newcite{uryupina2013multilingual} and \newcite{poesio2018crac}. 
The \newcite{uryupina2013multilingual} system takes all the {\NP}s as candidates and trains a SVM-based binary classifier to select mentions from all the {\NP}s. \newcite{poesio2018crac} briefly discuss %introduced 
a neural mention detector that they modified from the {\NER} system of \newcite{glample2016-ner}. The system uses a bidirectional {\LSTM} followed by a FFNN to select mentions from spans up to a maximum width. 
The system achieved 
substantial %large 
gains on mention F1 when compared with the \cite{lee-et-al:CL13} on {\CONLL} and {\CRAC} data sets. 

\textbf{Neural Named Entity Recognition.} %In fact, 
A subtask of %the 
mention detection that focuses only on detecting 
\textit{named} entity mentions has been studied more frequently. 
However, most of the proposed approaches treat the {\NER} task as a sequence labelling task,
thus cannot be directly applied in
tasks that require nested mentions, 
such as {\NER} in the biomedical domain or
coreference.
%the {\MD} task for coreference, as the later usually allow nested mentions. 
The first neural network based {\NER} model 
was %is
introduced by \newcite{collobert2011natural}, who used a {\CNN} to encode the tokens and applied a {\CRF} layer on top. After that, many other network architectures for {\NER} {\MD} 
have also been 
proposed,
such as {\LSTM}-{\CRF} \cite{glample2016-ner,chiu2016named}, {\LSTM}-{\CRF} + {\ELMO} \cite{peters2018elmo} and {\BERT} \cite{devlin2019bert}.

Recently, a number of {\NER} systems based on neural network architectures have been introduced to solve nested {\NER}. \newcite{ju-etal-2018-neural} introduce a stacked {\LSTM}-{\CRF} approach to solve nested {\NER} in multi-steps. \newcite{sohrab-miwa-2018-deep} use an exhaustive
region classification model. \newcite{lin-etal-2019-sequence} solve the problem in two steps: they first detect the entity head, and then infer the entity boundaries and classes in the second step. \newcite{strakova-etal-2019-neural} infer the nested {\NER} by a sequence-to-sequence model. \newcite{zheng2019boundary} introduce a boundary aware network to train the boundary detection and the entity classification models in a multi-task learning setting. However, none of those systems can be directly used for  coreference, due to the large difference between the settings used in {\NER} and in coreference (e.g. for coreference the mention need to be predicted in a {\HIGHRECALL} fashion). 
By contrast, our systems can be easily extended to do  nested {\NER}; we demonstrate this by evaluating our system on the {\GENIA} corpus.

\section{System architecture}

\begin{figure*}[t]
    \centering
    \includegraphics[width=\textwidth]{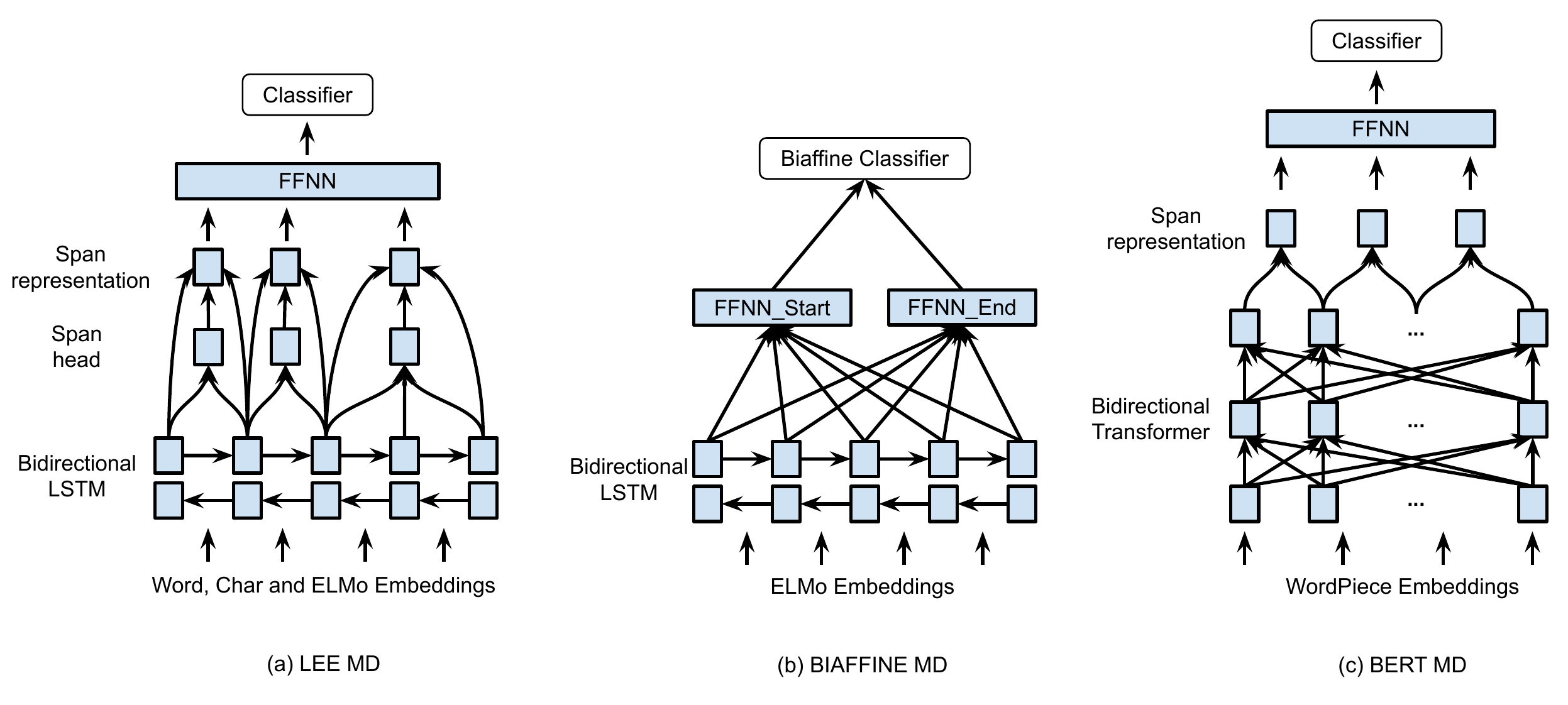}
    \caption{The overall network architectures of our approaches. (a) Our first approach that modified from \newcite{lee2018higher} coreference system. (b) Our second approach that uses biaffine classifier \cite{dozat-and-manning2017-parser}. (c) Our third approach that uses {\BERT} \cite{devlin2019bert} to encode the document.}
    \label{fig:nngraph}
\end{figure*}

Mention detection %for coreference 
is  the
task of extracting candidate mentions from the document. For a given document $D$ with $T$ tokens, we define all possible spans in $D$ as $N_{i=1}^{I}$ where $I = \frac{T(T+1)}{2}$, $s_i,e_i$ are the start and the end indices of $N_i$ where $1 \leq i \leq I$. 
The task for an {\MD} system is to assign all the spans ($N$) a score ($r_m$) so that spans can be classified into two classes (mention or non mention), hence is a binary classification problem. 

In this paper, we introduce three {\MD} systems\footnote{The code is available at 
\url{https://github.com/juntaoy/dali-md}}
that use the most recent neural network architectures.
The first approach uses the mention detection part from a start-of-the-art coreference resolution system \cite{lee2018higher}, 
which we refer to 
as %it 
{\LEEMD}. 
We remove the coreference part of the system and change the loss function to  sigmoid cross entropy, that is commonly used for binary classification problems. 
The second approach ({\BIAFFINEMD}) uses a bi-directional {\LSTM} to encode the sentences of the document, followed by a biaffine classifier \cite{dozat-and-manning2017-parser} to score the candidates. 
The third approach ({\BERTMD}) uses {\BERT} \cite{devlin2019bert} to encode the document in the sentence level; in addition, a feed-forward neural network (FFNN) to score the candidate mentions.
The three architectures are summarized in Figure  \ref{fig:nngraph} and discussed in detail below.
% Figure \ref{fig:nngraph} shows the overall network architectures of our three approaches.

All three architectures are available in two output modes:
{\HIGHF} and {\HIGHRECALL}.
%, we have two output modes ({\HIGHF} and {\HIGHRECALL}). 
The {\HIGHF} mode is meant for applications that require highest accuracy, such as preprocessing for annotation or nested {\NER}.
%The {\HIGHF} mode aims to achieve a better F1 score, which can be used by applications that require the best overall performance. 
The {\HIGHRECALL} mode, on the other hand, predicts as many mentions as possible, 
which is more appropriate for preprocessing for a coreference system since mentions can be further filtered by the system during coreference resolution. 
%The {\HIGHRECALL} mode, on the other hand, predicts more mentions than the {\HIGHF} mode by priorities the recall which fits the coreference system very well since mentions can be further filtered by the system during coreference resolution.
In {\HIGHF} mode we output mentions whose probability $p_m(i)$ is larger then a threshold $\beta$ such as 0.5. 
In {\HIGHRECALL} mode we output mentions based on a fixed mention/word ratio $\lambda$; this is the same method used by  \newcite{lee2018higher}.
%For {\HIGHF} mode we outputs mentions have a probability ($p_m(i)$) larger then a threshold ($\beta$) such as 0.5. For {\HIGHRECALL} mode we output mentions based on a fixed mention/word ratio ($\lambda$) which is the same method used in the \newcite{lee2018higher} system.

\subsection{\LEEMD}
Our first 
system is based on
%approach uses 
the mention detection part of the \newcite{lee2018higher} system. 
The system represents a candidate span with the outputs of a bi-directional {\LSTM}. 
The sentences of a document are encoded bidirectional via the {\LSTM}s to obtain forward/backward representations for each token in the sentence. 
The bi-directional {\LSTM} takes as input the concatenated embeddings ($(x_t)_{t=1}^{T}$) of both word and character levels.
For word embeddings,  GloVe \cite{pennington2014glove} and {\ELMO} \cite{peters2018elmo} embeddings are used. 
Character embeddings are learned from convolution neural networks ({\CNN}) during training. 
The tokens are represented by concatenated outputs from the forward and the backward {\LSTM}s. 
The token representations $(x^{*}_t)_{t=1}^{T}$  are used together with head representations ($h^*_i$) to represent candidate spans ($N^*_i$). The $h^*_i$ of a span is obtained by applying an attention over its token representations ($\{x^{*}_{s_i}, ..., x^{*}_{e_i}\}$), where $s_i$ and $e_i$ are the indices of the start and the end of the span respectively. Formally, we compute $h^*_i$, $N^*_i$ as follows:

$$\alpha_t = \textsc{FFNN}_{\alpha}(x^{*}_{t})$$
$$a_{i,t} = \frac{exp(\alpha_t)}{\sum^{e_i}_{k=s_i} exp(\alpha_k)}$$
$$h^*_i = \sum^{e_i}_{t=s_i} a_{i,t} \cdot x_t$$
$$N^*_i = [x^*_{s_i}, x^*_{e_i},h^*_i,\phi(i)]$$
where $\phi(i)$ is the span width feature embeddings.

To make the task computationally tractable, the model only considers the spans up to a maximum length of $l$, i.e. $e_i -  s_i < l, \ (s_i,e_i) \in N$. The span representations are passed to a FFNN to obtain the raw candidate scores ($r_m$). The raw scores are then used to create the probabilities ($p_m$) by applying a sigmoid function to the $r_m$:

$$r_m(i) = \textsc{FFNN}_{m} (N^*_i) $$
$$p_m(i) = \frac{1}{1+e^{-r_m(i)}}$$

For the {\HIGHRECALL} mode, the top ranked $\lambda T$ spans are selected from $lT$ candidate spans ($\lambda < l$) by ranking the spans in a descending order by their probability ($p_m$).
For the {\HIGHF} mode, the spans that have a probability ($p_m$) larger than the threshold $\beta$ are returned.

\subsection{\BIAFFINEMD}
In our second model,
%For our second approach, 
the same bi-directional {\LSTM} is used to encode the tokens of a document in the sentence level. 
However, instead of using the concatenations of multiple word/character embeddings, 
only {\ELMO} embeddings are used, 
as
%we use only the ELMo embeddings for our second approach. 
we find in %the 
preliminary experiments that the additional GloVe embeddings and character-based embeddings do not improve the accuracy.
%thus we discard them from our {\BIAFFINEMD}. 
After obtaining the token representations from the bidirectional {\LSTM}, we apply two separate %smaller 
FFNNs to create different representations ($h_s/h_e$) for the start/end of the spans. 
%By
Using different representations for the start/end of the spans allows the system to learn important information to identify the start/end of the spans separately. 
This 
is an advantage %has the conceptual advantage 
when compared to the model directly using the output states of the {\LSTM}, since the tokens that are likely to be the start of the mention and end of the mention are very different. 
Finally, we employ a biaffine attention \cite{dozat-and-manning2017-parser} over the sentence to create a $l_s \times l_s$ scoring metric ($r_m$), where $l_s$ is the length of the sentence. More precisely, we compute the raw score for span i ($N_i$) by:

$$h_{s}(i) = \textsc{FFNN}_{s}(x_{s_i}^*)$$
$$h_{e}(i) = \textsc{FFNN}_{e}(x_{e_i}^*)$$
$$r_m(i) = h_{s}(i)^\top W_{m} h_{e}(i) + h_{s}(i) b_{m}$$
where $s_i$ and $e_i$ are the start and end indices of $N_i$, $W_{m}$ is a $d \times d$ metric and $b_{m}$ is a bias term which has a shape of $d \times 1$. 

The computed raw score ($r_m$) covers all the span combinations in a sentence, to compute the probability scores ($p_m$) of the spans we further apply a simple constrain ($s_i \leq e_i$) such that the system only predict valid mentions. 
Formally:

$$p_m(i) = \Bigg\{
\begin{tabular}{ll}
     $\frac{1}{1+e^{-r_m(i)}}$& $s_i \leq e_i$  \\
     $0$ & $s_i > e_i$ 
\end{tabular}$$

The resulted $p_m$ are then used to predict mentions by filtering out the spans according to different requirements ({\HIGHRECALL} or {\HIGHF}).

\subsection{\BERTMD}
Our third approach is based on the recently introduced {\BERT} model
\cite{devlin2019bert} 
in which  sentences are encoded using deep bidirectional transformers. 
Our model uses a pre-trained {\BERT} model to encode the documents in the sentence level to create token representations $x_t^*$. The pre-trained {\BERT} model uses WordPiece embeddings \cite{wu2016google}, in which tokens are further split into smaller word pieces as the name suggested. For example in sentence:

\textit{We respect \#\#fully invite you to watch a special edition of Across China .}

The token ``respectfully" is split into two pieces (``respect" and ``fully"). 
If %In the case that 
tokens have multiple representations (word pieces), we use the first representation of the token. An indicator list is created during the data preparation step to link the tokens to the correct word pieces. After obtaining the actual word representations, the model then creates candidate spans by considering spans up to a maximum span length ($l$). The spans are represented by the concatenated representations of the start/end tokens of the spans. This is followed by a FFNN and a sigmoid function to assign each span a probability score:

$$N_i^* = [x_{s_i}^*, x_{e_i}^*]$$
$$r_m(i) = \textsc{FFNN}_{m} (N^*_i) $$
$$p_m(i) = \frac{1}{1+e^{-r_m(i)}}$$

We use the same methods we used for our first approach ({\LEEMD}) to select mentions based on different settings ({\HIGHRECALL} or {\HIGHF}) respectively.

\subsection{Nested {\NER}} In our evaluation on nested {\NER} we assign each mention pairs $C$ raw scores ($r_m$) ($C= 1+ $ number of {\NER} categories). The first score indicates the likelihood of a span is \textit{not} a named mentions, the rest of the scores are corresponding to individual {\NER} categories. The probability ($p_m$) is then calculated by the softmax function instead of the sigmoid function:

$$p_m(i_c) = \frac{e^{r_m(i_c)}}{\sum^{C}_{\hat{c}=1} e^{r_m(i_{\hat{c}})}}$$

\subsection{Learning}
The learning objective of our mention detectors is to learn to distinguish mentions form non-mentions. Hence it is a binary classification problem, we optimise our models on the sigmoid cross entropy. 

$$- \sum_{i=1}^N y_i \log p_m(i) + (1-y_i) \log (1-p_m(i))$$
where $y_i$ is the gold label ($y_i \in \{0,1\}$) of i$_{th}$ spans.

For our further experiments on the nested {\NER} we use the softmax cross entropy instead:

$$- \sum_{i=1}^N\sum_{c=1}^C y_{i_c} \log p_m(i_c)$$

\section{Experiments}

We ran three series of experiments.
%Our experiments consist of two major parts. 
The first series of experiments
%The first part of the experiment 
focuses only on the mention detection task, and we evaluate the performance of the proposed mention detectors in isolation.
The second series of experiments
%The second part of the experiment 
focuses on the effects of our model on 
coreference: %the downstream applications: 
i.e., 
we integrate the mentions extracted from our best system into state-of-the-art coreference systems (both end-to-end and the pipeline system). 
The third series of experiments focuses on the nested {\NER} task. 
We evaluate our systems  both on boundary detection and on the full {\NER} tasks.
The 
rest %remaining part 
of this section introduces our experimental settings in detail.

\subsection{Data Set}
We evaluate our models on two different corpora for both the mention detection and the coreference tasks and one additional corpora for nested {\NER} task, the {\CONLL} 2012 English corpora \cite{pradhan2012conllst}, the {\CRAC} 2018 corpora \cite{poesio2018crac} and the {\GENIA} \cite{kim2003genia} corpora. 

The {\CONLL} data set
is the standard reference corpora for coreference resolution. 
The English subset consists of 2802, 342, and 348 documents for the train, development and test sets respectively. 
The {\CONLL} data set is not however ideal for mention detection, since not all  mentions are annotated, 
but only mentions involved in coreference chains of length $> 1$. %i.e. they did not annotate the singleton mentions. 
This has a negative impact on learning since 
%the 
singleton mentions will always receive negative labels. 

The {\CRAC} corpus uses data from the {\ARRAU} corpus \cite{uryupina-et-al:NLEJ}.
{\ARRAU} consists of texts from four very distinct domains: news (the \ACRO{rst} subcorpus), dialogue (the \ACRO{trains} subcorpus) and fiction (the \ACRO{pear} stories). 
This corpus is more appropriate for studying mention detection as all  mentions are annotated.
As done in the {\CRAC} shared task, we used the \ACRO{rst} portion of the corpora, consisting of news texts (1/3 of the \ACRO{penn} Treebank). 
Since none of the state-of-the-art coreference systems predict singleton mentions, a  version of the {\CRAC} dataset with  singleton mentions excluded was created for the coreference task evaluation. 

The {\GENIA} corpora is one of the main resources for studying nested {\NER}. 
We use the {\GENIA} v3.0.2 corpus and preprocess the dataset following the same settings of \newcite{finkel-manning-2009-nested} and \newcite{lu-roth-2015-joint}. 
Historically, the dataset has been split into two different ways: the first approach splits the data into two sets (train and test) by 90:10 ({\GENIATT}),
whereas the second approach further creates a development set by splitting the data into 81:9:10 ({\GENIATDT}). We evaluate our model on both approaches to make the fair comparisons with previous work. For evaluation on {\GENIATT}, since we do \textit{not} have a development set, we train our model for 40K steps (20 epochs) and take evaluate on the final model.

\subsection{Evaluation Metrics}
For our experiments on the mention detection or nested {\NER}, we report recall, precision and F1 scores for mentions. For our evaluation that involves the coreference system, we use the official {\CONLL} 2012 scoring script to score our predictions.
Following standard practice, we report recall, precision, and F1 scores for MUC, B$^3$ and CEAF$_{\phi_4}$ and the average F1 score of those three metrics. 

\subsection{Baseline System}
For the mention detection evaluation %task 
we use the \newcite{lee2018higher} system as baseline. The baseline is trained end-to-end on the coreference task and we use as baseline the mentions predicted by the system before carrying out coreference resolution. 

For the coreference evaluation %task 
we use the top performing \newcite{lee2018higher} system as our baseline for the end-to-end system, and the \newcite{clark2016deep} system as our baseline for the pipeline system. 
During the evaluation, we slightly modified the \newcite{lee2018higher} system to allow the system to take the mentions predicted by our model instead of its internal mention detector. Other than that we keep the system unchanged.

For the nested {\NER} we compare our system with \newcite{zheng2019boundary} and \newcite{sohrab-miwa-2018-deep} on {\GENIATDT} and with \newcite{ju-etal-2018-neural}, \newcite{lin-etal-2019-sequence} an \newcite{strakova-etal-2019-neural} on {\GENIATT}.

\subsection{Hyperparameters}

\begin{table}[t]
    \centering
    \begin{tabular}{l l l}
    \toprule
    \bf Model& \bf Parameter & \bf Value \\
    \midrule
    {\TABLEEMD}, {\TABBIAFFINEMD} &BiLSTM layers &3\\
    {\TABLEEMD}, {\TABBIAFFINEMD} &BiLSTM size &200\\
    {\TABLEEMD}, {\TABBIAFFINEMD} &BiLSTM dropout &0.4\\
    {\TABBERTMD} &Transformer layers & 12\\
    {\TABBERTMD} &Transformer size & 768\\
    {\TABBERTMD} &Transformer dropout & 0.1\\
    {\TABLEEMD}, {\TABBIAFFINEMD}, {\TABBERTMD}& FFNN layers & 2\\
    {\TABLEEMD}, {\TABBIAFFINEMD}, {\TABBERTMD}& FFNN size & 150\\
    {\TABLEEMD}, {\TABBIAFFINEMD}, {\TABBERTMD}& FFNN dropout & 0.2\\
    {\TABLEEMD}, {\TABBIAFFINEMD}, {\TABBERTMD}& Embeddings dropout & 0.5\\
    {\TABLEEMD}, {\TABBIAFFINEMD}, {\TABBERTMD}& Optimiser & Adam\\
    {\TABLEEMD}, {\TABBIAFFINEMD}& Learning rate & 1e-3\\
    {\TABBERTMD} & Learning rate & 2e-5\\
    {\TABLEEMD}, {\TABBIAFFINEMD}, {\TABBERTMD}& Training step & 40K\\
    \bottomrule
    \end{tabular}
    \caption{Major hyperparameters for our models. {\TABLEEMD}, {\TABBIAFFINEMD}, {\TABBERTMD} are used to indicate {\LEEMD}, {\BIAFFINEMD}, {\BERTMD} respectively.}
    \label{tab:hyperparameters}
\end{table}

For our first model ({\LEEMD}) we use the default settings of \newcite{lee2018higher}. For word embeddings the system uses 300-dimensional GloVe embeddings \cite{pennington2014glove} and 1024-dimensional {\ELMO} embeddings \cite{peters2018elmo}. The character-based embeddings are produced by a convolution neural network (\CNN) which has a window sizes of 3, 4, and 5 characters (each has 50 filters). The characters embeddings (8-dimensional) are randomly initialised and learned during the training. The maximum span width is set to 30 tokens. 

For our {\BIAFFINEMD} model, we use the same {\LSTM} settings and the hidden size of the FFNN as our first approach. For word embeddings, we only use the {\ELMO} embeddings \cite{peters2018elmo}. 

For our third model ({\BERTMD}), we fine-tune on the pre-trained {\BERT}$_{\textsc{base}}$ that consists of 12 layers of transformers. The transformers use 768-dimensional hidden states and 12 self-attention heads. The WordPiece embeddings \cite{wu2016google} have a vocabulary of 30,000 tokens. We use the same maximum span width as in our first approach (30 tokens). 

The detailed network settings can be found in Table \ref{tab:hyperparameters}.

\section{Results and Discussions}

In this section, we first evaluate the proposed models
in isolation
%by themselves 
on the mention detection task. 
We then integrate the mentions predicted by our system into coreference resolution systems to evaluate the effects of our {\MD} systems on the downstream applications. 
Finally we evaluate our system on the nested {\NER} task.

\subsection{Mention Detection Task}
\textbf{Evaluation on the {\CONLL} data set.} 
For mention detection 
%task 
%evaluated 
on the {\CONLL} data set, we first take the best model from \newcite{lee2018higher} and use its default mention/token ratio ($\lambda = 0.4$) to output predicted mentions before coreference resolution. We use this as our baseline for the {\HIGHRECALL} setting. We then evaluate all three proposed models with the same $\lambda$ as that of the baseline. As a result, the number of mentions predicted by different systems is the same, which means 
%the 
mention precision will be similar. 
Thus, for the {\HIGHRECALL} setting we compare the systems by %the 
mention recall.  
As we can see from 
Table
%table 
\ref{tab:emdhighrecall}, the baseline system already achieved a reasonably good recall of 96.6\%. 
But even when compared with such a strong baseline,
%Start from a strong baseline, 
by simply separately training the mention detection part of the baseline system, the stand-alone {\LEEMD} achieved an improvement of 0.7 p.p. 
This indicates that mention detection task does not benefit from  joint mention detection and coreference resolution. The {\BERTMD} achieved 
the same recall
%a recall same 
as the {\LEEMD},
but {\BERTMD} uses 
a much deeper network and is more expensive to train. 
By contrast, the {\BIAFFINEMD} uses the simplest network architecture 
among %amount 
the three approaches,  yet achieved the best results, 
outperforming %which outperforms 
the baseline by 0.9 p.p. (26.5\% error reduction).

\textbf{Evaluation on the {\CRAC} data set\footnote{As the \newcite{lee2018higher} system does not predict singleton mentions, the results on {\CRAC} data set in Table \ref{tab:emdhighrecall} are evaluated without singleton mentions, whereas the results reported in Table \ref{tab:emdfinal} are evaluated with singleton mentions included.} }
For %experiment on 
the {\CRAC} data set, we train the \newcite{lee2018higher} system end-to-end on the reduced corpus 
with %that have 
singleton mentions removed and extract mentions from the system by set $\lambda = 0.4$. 
We then train our models with the same $\lambda$ but on the full corpus, since our mention detectors naturally support both mention types (singleton and non-singleton mentions). Again, the baseline system has a decent recall of 95.4\%. 
Benefiting %Benefit 
from the singletons, our {\LEEMD} and {\BIAFFINEMD} 
models achieved larger improvements when compared with the gains achieved on the {\CONLL} data set. The largest improvement  (1.8 p.p.) is achieved by our {\BIAFFINEMD} model with an error reduction rate of 39.1\%. 
%The 
{\BERTMD} achieved a relatively smaller gain (0.8 p.p.) when compared with the other models;
this might be a result of the difference in corpus size %different 
between {\CRAC} and {\CONLL} data set. 
(The {\CRAC} corpus is smaller than the {\CONLL} data set.)

\begin{table}[t]
    \centering
    \begin{tabular}{lllll}
    \toprule
   \bf Data&\bf Model & \bf R &\bf P &\bf F1 \\
   \midrule
    \multirow{4}{*}{\CONLL}&
    \newcite{lee2018higher}&96.6&28.2&43.7\\ \cmidrule{2-5}
    &\LEEMD&97.3&28.4&44.0\\
    &\BIAFFINEMD&\bf 97.5&\bf 28.5&\bf 44.1\\
    &\BERTMD&97.3&28.4&44.0\\
    \midrule
    \multirow{4}{*}{{\CRAC}$^3$}&
    \newcite{lee2018higher}&95.4 &34.4 &50.6\\ \cmidrule{2-5}
    &\LEEMD&96.9 &\bf 35.0 &51.4\\
    &\BIAFFINEMD&\bf 97.2 &\bf 35.0 &\bf 51.5\\
    &\BERTMD&96.2 &34.7& 51.0\\
    \bottomrule
    \end{tabular}
    \caption{Performance comparison between our mention detectors and the baseline (\newcite{lee2018higher} system) in a {\HIGHRECALL} setting.}
    \label{tab:emdhighrecall}
\end{table}

\begin{table}[t]
    \centering
    \begin{tabular}{lllll}
    \toprule
   \bf Data&\bf Model & \bf R &\bf P &\bf F1 \\
   \midrule
    \multirow{5}{*}{\CONLL}&
     \newcite{lee-et-al:CL13}&89.5&40.4&55.7\\
    &{\newcite{poesio2018crac}}&74.0&73.5&73.8\\\cmidrule{2-5}
    &\HIGHRECALL&\bf 97.5&28.5&44.1\\
    &\HIGHF&76.0&\bf 82.6&\bf 79.1\\
    &\BALANCE&90.6&53.0&66.9\\
    \midrule
    \multirow{4}{*}{{\CRAC}$^3$}&
     \newcite{lee-et-al:CL13}&67.3& 71.6& 69.4\\
    &{\newcite{poesio2018crac}}&86.2& 79.3& 82.6\\\cmidrule{2-5}
    &\HIGHRECALL&\bf 95.3& 74.2& 83.5\\
    &\HIGHF&87.9& \bf 89.7& \bf 88.8\\
    \bottomrule
    \end{tabular}
    \caption{Comparison between our {\BIAFFINEMD} and the top performing systems on the mention detection task using the {\CONLL} and {\CRAC} data sets.}
    \label{tab:emdfinal}
\end{table}

\begin{table}[t]
    \centering
    \begin{tabular}{lrcc}
    \toprule
    & \bf Num of &\multicolumn{2}{c}{\bf Infer Speed}\\ 
   \bf Model & \bf Parameters &\bf {\CONLL}&\bf {\CRAC}\\
   \midrule
    \LEEMD&4 M&6.8&5.5\\
    \BIAFFINEMD&3.4 M&8.3&6.7\\
    \BERTMD&110 M&3.3&4.3\\
    \bottomrule
    \end{tabular}
    \caption{Model complexity and inference speed (docs/s) comparison between our mention detectors}
    \label{tab:speed}
\end{table}

\begin{table*}[t]
\centering
\begin{tabular}{l l l l l l l l l l l l}
\toprule
\multirow{2}{*}{\bf Data} & \multirow{2}{*}{\bf Model} & \multicolumn{3}{l}{\bf MUC} & \multicolumn{3}{l}{\bf B$^3$} & \multicolumn{3}{l}{\bf CEAF$_{\phi_4}$} & \multirow{2}{*}{\begin{tabular}[c]{@{}l@{}}\bf Avg.\\ \bf F1\end{tabular}} \\ \cmidrule{3-11}
 & & \bf P & \bf R & \bf F1& \bf P & \bf R & \bf F1& \bf P& \bf R  & \bf F1 & \\ \midrule
\multirow{8}{*}{\CONLL}
&\newcite{lee2018higher}&\bf 81.4 &\bf 79.5 & \bf 80.4 & \bf 72.2 &\bf 69.5 &\bf 70.8 &\bf 68.2 &\bf 67.1 & \bf 67.6 & \bf 73.0\\
&\ \ + {\HIGHRECALL} &80.0&\bf 79.5 &79.7 &70.5 &\bf 69.5 &70.0 &67.3 &66.9 &67.1 & 72.3\\ 
&\ \ + {\HIGHRECALL} + joint&80.9& 79.2&80.0 &72.0 &69.0 &70.5 &67.7 &66.9 &67.3 &72.6 \\ \cmidrule{3-12}
&\newcite{lee2017end}&78.4& 73.4& 75.8& 68.6& 61.8& 65.0& 62.7& 59.0& 60.8& 67.2\\ 
&\ \ + {\HIGHRECALL} &\bf 78.6&\bf 74.0 &\bf 76.2 &\bf  68.9&\bf  62.2&\bf  65.4&\bf 63.2 &\bf 59.6&\bf 61.4 & \bf 67.7\\ \cmidrule{3-12}
&\newcite{clark2016deep}&79.2& 70.4& 74.6& 69.9& 58.0& 63.4& 63.5& 55.5& 59.2& 65.7\\
&\ \ + {\HIGHRECALL} &78.7 & 72.4 & 75.4 & 69.4 & 59.7 & 64.2 & 62.2 & \bf 57.7 & 59.9 & 66.5 \\ 
&\ \ + {\BALANCE} &\bf 80.3 & \bf 72.5 & \bf 76.2 & \bf 71.2 & \bf 60.4 & \bf 65.3 & \bf 64.6 & 57.1 & \bf 60.6 & \bf 67.4 \\ 
 \midrule
\multirow{3}{*}{\CRAC}
&\newcite{lee2018higher}&\bf 79.2&71.9&75.3&\bf 72.4&63.5&67.7&66.2&58.6&62.2&68.4\\
&\ \ + {\HIGHRECALL} &76.2&73.1&74.6 &68.4 & \bf 65.5&66.9 &65.1 &61.8 &63.4 & 68.3\\ 
&\ \ + {\HIGHRECALL} + joint&77.6&\bf 73.4 &\bf 75.4 &70.4 &\bf 65.5 &\bf 67.9 &\bf 66.4 &\bf 61.9 &\bf 64.1 &\bf 69.1 \\ 
\bottomrule
\end{tabular}

\caption{\label{tab:coref} Comparison between the baselines and the models enhanced by our {\BIAFFINEMD} on the coreference resolution task.}
\end{table*}

\textbf{Model complexity and speed} To give an idea of the difference in model complexity and  inference speed, we listed the number of trainable parameters and the inference speed of our models on {\CONLL} and {\CRAC} in Table \ref{tab:speed}.
The {\BIAFFINEMD} model consists of the lowest number of trainable parameters among all three models, it uses 85\% or 3\% parameters when compared with the {\LEEMD} and {\BERTMD} respectively.  In addition, the {\BIAFFINEMD} is also the fastest model on both {\CONLL} and {\CRAC} datasets, which is able to process 8.3 {\CONLL} or 6.7 {\CRAC} documents per second\footnote{All the speed test is conducted on a single GTX 1080ti GPU.}.

\textbf{Comparison with the State-of-the-art.} We compare our best system {\BIAFFINEMD} with the rule-based mention detector of the Stanford deterministic system \cite{lee-et-al:CL13} and the neural mention detector of \newcite{poesio2018crac}. For {\HIGHF} setting we use the common threshold ($\beta = 0.5$) for binary classification problems without tuning. 
For evaluation on {\CONLL} we 
create in addition
%in additional create 
a variant %variance 
of the {\HIGHRECALL} setting ({\BALANCE}) by setting $\lambda=0.2$; 
this is 
because 
%due to 
we noticed that the score differences between the {\HIGHRECALL} and {\HIGHF} settings are relatively large (see Table \ref{tab:emdfinal}). 
The score differences between our two settings on {\CRAC} data set are smaller; this might because the {\CRAC} data set annotated both singleton and non-singleton mentions, hence the models are trained in a more balanced way.  
Overall, when compared with the best-reported system \cite{poesio2018crac}, our {\HIGHF} settings outperforms their system by large margin of 5.3\% and 6.2\% on {\CONLL} and {\CRAC} data sets respectively.

\subsection{Coreference Resolution Task}
We then integrate the mentions predicted by our best system into the coreference resolution system to evaluate the effects of our better mention detectors on the downstream application. 

\textbf{Evaluation with an end-to-end system.} We first evaluate our {\BIAFFINEMD} 
in combination with %on 
the end-to-end \newcite{lee2018higher} system. We slightly modified the system to feed the system mentions predicted by our mention detector. As a result, the original mention selection function is switched off, we keep all the other settings (include the mention scoring function) unchanged. We then train the modified system to obtain a new model. As illustrated in Table \ref{tab:coref}, the model trained using mentions supplied by our {\BIAFFINEMD} achieved a F1 score slightly lower than the original end-to-end system, 
even though %dispute 
our mention detector has a better performance. 

We 
think %suggest 
the performance drop might be the result of two 
factors. %reasons. 
First, by replacing the original mention selection function, the system actually becomes a pipeline system, thus cannot benefit from joint learning. 
Second, the performance difference between our mention detector and the original mention selection function might not be large enough to deliver improvements on the final coreference results. 
To test %confirm 
our hypotheses, we evaluated our {\BIAFFINEMD} with two additional experiments. 

In the first experiment, we enabled the original mention selection function and fed the system slightly more mentions. More precisely, we configured our {\BIAFFINEMD} to output 0.5 mention per token instead of 0.4 i.e. $\lambda=0.5$. As a result, the coreference system has the freedom to select its own mentions from a candidate pool supplied by our {\BIAFFINEMD}. After training the system with the new setting, we get an average F1 of 72.6\% (see table \ref{tab:coref}), which narrows the performance gap between the end-to-end system and the model trained without the joint learning. This confirms our first hypothesis that by downgrading the system to a pipeline setting does harm the overall performance of the coreference resolution.

For our second experiment, we used the \newcite{lee2017end} instead. The \newcite{lee2018higher} system is an extended version of the \newcite{lee2017end} system, hence they share most of the network architecture. The \newcite{lee2017end} has a lower performance on mention detection (93.5\% recall when $\lambda=0.4$), which creates a large (4\%) difference when compared with the recall of our {\BIAFFINEMD}. We train the system without the joint learning, and the newly trained model achieved an average F1 of 67.7\% and this is 0.5 better than the original end-to-end \newcite{lee2017end} system (see table \ref{tab:coref}). This confirms our second hypothesis that a larger gain on mention recall is needed in order to show improvement on the overall system. 

We further evaluated the \newcite{lee2018higher} system on the {\CRAC} data set. 
We first train the original \newcite{lee2018higher} on the reduced version (with singletons removed) of the {\CRAC} data set to create a baseline. 
As we can see from Table \ref{tab:coref}, the baseline system has an average F1 score of 68.4\%. We then evaluate the system with mentions predicted by our {\BIAFFINEMD}, we experiment with both joint learning disabled and enabled. 
As shown %showed 
in Table %table 
\ref{tab:coref}, the model without joint learning achieved an overall score 0.1\% lower than the baseline, but the new model has clearly a better recall on all three metrics when compared with the baseline. The model trained with joint learning enabled achieved an average F1 of 69.1\% which is 0.7\% better than the baseline.

\textbf{Evaluation with a pipeline system.} We then evaluated our best model ({\BIAFFINEMD}) with a pipeline system. We use the best-reported pipeline system by \newcite{clark2016deep} as our baseline. 
The original system used the rule-based mention detector from the Stanford deterministic coreference system \cite{lee-et-al:CL13} (a performance comparison between the \newcite{lee-et-al:CL13} and our {\BIAFFINEMD} can be found in Table \ref{tab:emdfinal}). 
We modified the preprocessing pipeline of the system to use mentions predicted by our {\BIAFFINEMD}. 
We ran the system with both mentions from the {\HIGHRECALL} and {\BALANCE} settings, as both settings have reasonable good mention recall which is required to train a coreference system. After training the system with mentions from our {\BIAFFINEMD}, the newly obtained models achieved large improvements of 0.8\% and 1.7\% for {\HIGHRECALL} and {\BALANCE} settings respectively. This suggests that the \newcite{clark2016deep} system works better on a smaller number of high-quality mentions than a larger number but lower quality mentions. We also noticed that the speed of the \newcite{clark2016deep} system is sensitive to the size of the predicted mentions, both training and testing finished much faster when tested on the {\BALANCE} setting. We did not test the \newcite{clark2016deep} system on the {\CRAC} data set, as a lot of effects are needed to fulfil the requirements of the preprocessing pipeline, e.g. predicted parse trees, named entity tags.  Overall our {\BIAFFINEMD} showed its merit on enhancing the pipeline system.

\textbf{Summary}
In summary, our results suggest that the picture regarding using our {\BIAFFINEMD} for coreference \textit{resolution} is more mixed than with coreference \textit{annotation} as discussed in the previous Section (and with nested {\NER} as shown in the following Section). 
Our model clearly improves the results of best current pipeline system; 
when used with a top performing end-to-end system, it improves the performance with the {\CRAC} dataset but not with {\CONLL}.

\subsection{Nested Named Entity Recognition Task}
We then extend our best system ({\BIAFFINEMD}) to do nested {\NER} task.

\begin{table}[t]
    \centering
    \begin{tabular}{llll}
    \toprule
   \bf Model & \bf R &\bf P&\bf F1\\
   \midrule
    \newcite{zheng2019boundary}&73.6&75.9&74.7\\
    \newcite{sohrab-miwa-2018-deep}&64.0&\bf 93.2&77.1\\\midrule
    Our model&\bf 78.5&77.8&\bf 78.2\\
    \bottomrule
    \end{tabular}
    \caption{Overall performance comparison on {\GENIATDT} test set.}
    \label{tab:genia81_overall}
\end{table}

\begin{table}[t]
    \centering
    \begin{tabular}{llll|l|l}
    \toprule
   \bf Categories & \bf R &\bf P&\bf F1&\bf So F1 &\bf Zh F1 \\
   \midrule
    DNA&74.0&75.3&\bf 74.6&71.8&70.6\\
    RNA&85.3&84.5&\bf 84.9&72.4&81.5\\
    protein&82.6&78.5& 80.5&\bf 80.8&76.4\\
    cell line&70.1&77.4&\bf 73.6&67.9&71.3\\
    cell type&72.1&78.3&75.1&\bf 78.1&72.5\\
    \bottomrule
    \end{tabular}
    \caption{Individual category performance comparison on {\GENIATDT} test set. Zh refers to \newcite{zheng2019boundary} and So refers to \newcite{sohrab-miwa-2018-deep}}
    \label{tab:genia81_category}
\end{table}

\textbf{Evaluation on {\GENIATDT}} We first evaluate our system on the split ({\GENIATDT}) with the development set. We first run our {\BIAFFINEMD} on boundary detection task which do not require any modification on the system. On boundary detection our system achieved 80.0\%, 82.3\% and 81.1\% for recall precision and F1 score respectively. Our results out perform the previous state-of-the-art system \cite{zheng2019boundary} by 2.8\% (F1 score). 

We then extend our system to predict full {\NER} task, Table \ref{tab:genia81_overall} and Table \ref{tab:genia81_category} show our overall and individual category results on the {\GENIATDT} test set respectively. As we can see from Table \ref{tab:genia81_overall} our system outperforms the previous state-of-the-art system by 1.1 percentage points. In addition, our system also achieved a much better results on three out of five categories (see Table \ref{tab:genia81_category}). Overall our system achieved the new state-of-the-art on the {\GENIATDT} data for both boundary detection and full nested {\NER} tasks.

\begin{table}[t]
    \centering
    \begin{tabular}{llll}
    \toprule
   \bf Model & \bf R &\bf P&\bf F1\\
   \midrule
   \newcite{ju-etal-2018-neural}&71.3&78.5&74.7\\
    \newcite{lin-etal-2019-sequence}&73.9&75.8&74.8\\
    \newcite{strakova-etal-2019-neural}&-&-&78.3\\\midrule
    Our model&\bf 78.0&\bf 79.1&\bf 78.6\\
    \bottomrule
    \end{tabular}
    \caption{Overall performance comparison on {\GENIATT} test set.}
    \label{tab:genia90_overall}
\end{table}

\begin{table}[t]
    \centering
    \begin{tabular}{llll|l}
    \toprule
   \bf Categories & \bf R &\bf P&\bf F1&\bf Ju F1 \\
   \midrule
    DNA&72.4&78.9&\bf 75.5&72.0\\
    RNA&88.1&86.5&\bf 87.3&84.5\\
    protein&82.2&79.5&\bf 80.8&76.7\\
    cell line&67.4&80.0&\bf 73.2&71.2\\
    cell type&74.4&75.4&\bf 74.9&72.0\\
    \bottomrule
    \end{tabular}
    \caption{Individual category performance comparison on {\GENIATT} test set. Ju refers to \newcite{ju-etal-2018-neural} }
    \label{tab:genia90_category}
\end{table}

\textbf{Evaluation on {\GENIATT}} Next, we evaluate our system on the other split of the corpora ({\GENIATT}), in this setting we have a larger training set but do \textit{not} have a development set. After we train our model for 20 epochs, our final model outperforms the previous state-of-the-art by 0.3\% (see Table \ref{tab:genia90_overall}). In Table \ref{tab:genia90_category} we present our detailed scores for individual categories, since both \newcite{lin-etal-2019-sequence} and \newcite{strakova-etal-2019-neural} did not report the detailed scores, we compare our system with \newcite{ju-etal-2018-neural}. Our system outperforms theirs on all five categories.

\section{Conclusions}
In this work, we compare three neural network based approaches for mention detection. %task. 
The first model is 
a modified version of %modified from 
the mention detection part of a top performing coreference resolution system \cite{lee2018higher}. 
The second model used %the 
{\ELMO} embeddings together with a bidirectional {\LSTM}, and with a biaffine classifier on top. 
The third model adapted the {\BERT} model that based on the deep transformers and followed by a FFNN. 
We assessed the performance %robustness 
of our models in mention detection, coreference and nested {\NER} tasks. In the evaluation of mention detection, our proposed models reduced up to 26\% and 39\% of the recall error when compared with the strong baseline on {\CONLL} and {\CRAC} data sets in a {\HIGHRECALL} setting. The same model ({\BIAFFINEMD}) outperforms the best performing system on the {\CONLL} and {\CRAC} by large 5-6\% in a {\HIGHF} setting. In term of the evaluation on coreference resolution task,  by integrating our mention detector with the state-of-the-art coreference systems, we improved the end-to-end and pipeline systems by up to 0.7\% and 1.7\% respectively. The evaluation on the nested {\NER} task showed that despite our model is not specifically designed for the task, we achieved the state-of-the-art on the {\GENIA} corpora.
Overall, we introduced three neural mention detectors and showed that the improvements achieved on the mention detection task can be transferred to the downstream coreference resolution task. In addition, our model is robust enough be used for nested {\NER} task.

\section{Acknowledgments}

This research was supported in part by the DALI project, ERC Grant 695662.
\newpage

% \nocite{*}
\section{Bibliographical References}
\label{main:ref}

\bibliographystyle{lrec}
\bibliography{lrec2020}

%\section{Language Resource References}
%\label{lr:ref}
%\bibliographystylelanguageresource{lrec}
%\bibliographylanguageresource{lrec2020W-xample}

\end{document}